\documentclass[letterpaper, 10 pt, conference]{ieeeconf}  

\IEEEoverridecommandlockouts                              

\usepackage[pdftex]{graphicx}
\usepackage{amsmath}
\usepackage{amsfonts,amssymb}
\usepackage{bbm}

\usepackage{amsthm} 
\usepackage{thmtools}
\usepackage{mathtools}
\usepackage{epstopdf}
\usepackage{lipsum}  
\usepackage{xspace}

\usepackage[font=footnotesize]{caption}
\usepackage[font+=small]{subcaption} 

\usepackage{units}
\usepackage{booktabs} 
\usepackage{tabulary}
\usepackage[usenames,dvipsnames]{color} 

\usepackage{algorithm}
\usepackage{algpseudocode}
\algtext*{EndWhile}
\algtext*{EndIf}
\algtext*{EndFor}

\algblock[Input]{Input}{EndInput}
\algtext*{EndInput}
\algblock[Output]{Output}{EndOutput}
\algtext*{EndOutput}
\algblock[Variables]{Variables}{EndVariables}
\algtext*{EndVariables}

\usepackage[utf8]{inputenc}

\usepackage{pgfplotstable}
\pgfplotsset{
    tick label style={font=\small},
    label style={font=\small},
    legend style={font=\footnotesize}
}
\pgfplotsset{every axis plot/.append style={
        line width=1.5pt}
}

\date{}

\usepackage{ifthen,version}
\usepackage{color}
\usepackage{soul}
\usepackage{fixltx2e}
\usepackage[pdftex,colorlinks,bookmarks=true]{hyperref} 

\makeatletter
\def\thm@space@setup{\thm@preskip=0pt
\thm@postskip=0pt}
\makeatother

\newtheorem{theorem}{Theorem}
\newtheorem{lemma}{Lemma}
\newtheorem{problem}{Problem}
\newtheorem{definition}{Definition}

\newcommand{\fullFigGap}[0]{\vspace{-1.5\baselineskip}} 

\DeclareMathOperator*{\argmin}{arg\,min}
\DeclareMathOperator*{\argmax}{arg\,max}

\newcommand{\argmaxprob}[1]{\argmax\limits_{#1}}

\newcommand{\abs}[1]{\left|#1 \right|}
\newcommand{\card}[1]{\left|#1\right|}

\DeclareMathOperator*{\suchthat}{\;\; \mbox{s.t.} \;\;}
\newcommand{\expect}[2]{\mathbb{E}_{#1}\left[#2\right]}

\newcommand{\real}[0]{\mathbb{R}}

\newcommand{\bbm}{\begin{bmatrix}}
\newcommand{\ebm}{\end{bmatrix}}

\newcommand{\pair}[2]{\left( #1, #2\right)}

\newcommand{\seq}[2]{\left(#1_{1}, #1_{2}, \ldots, #1_{#2}\right)}

\newcommand{\setst}[2]{\left\lbrace #1\;\;\middle|\;\;#2\right\rbrace}

\newcommand{\vertexSet}[0]{\mathcal{V}}
\newcommand{\vertex}[0]{v}
\newcommand{\vertexGroup}[0]{V}

\newcommand{\vertexStart}[0]{\vertex_s}
\newcommand{\Path}[0]{\xi}
\newcommand{\PathSet}[0]{\Xi}

\newcommand{\world}[0]{\phi}

\newcommand{\worldSet}[0]{\mathcal{M}}
\newcommand{\meas}[0]{y}
\newcommand{\measSet}[0]{\mathcal{Y}}
\newcommand{\measFnDef}[0]{\mathcal{H}}
\newcommand{\measFn}[2]{\measFnDef\left(#1, #2\right)}

\newcommand{\utilityFnDef}[0]{\mathcal{F}}
\newcommand{\utilityFn}[2]{\utilityFnDef\left(#1, #2\right)}
\newcommand{\marginalGain}[2]{\Delta_\utilityFnDef\left(#1, #2\right)}

\newcommand{\costFnDef}[0]{\mathcal{T}}
\newcommand{\costFn}[2]{\costFnDef\left(#1, #2\right)}
\newcommand{\costBudget}[0]{B}

\newcommand{\state}[0]{s}
\newcommand{\stateSet}[0]{\mathcal{S}}
\newcommand{\action}[0]{a}

\newcommand{\actionSet}[0]{\mathcal{A}}
\newcommand{\actionSetFeas}[2]{\actionSet_{\mathrm{feas}}\left(#1, #2\right)}
\newcommand{\transFnDef}[0]{\Omega}
\newcommand{\transFn}[3]{\transFnDef{}\left(#1, #2, #3\right)}
\newcommand{\rewardFnDef}[0]{R}
\newcommand{\rewardFn}[3]{\rewardFnDef{}\left(#1, #2, #3\right)}

\newcommand{\policy}[0]{\pi}
\newcommand{\policySet}[0]{\Pi}

\newcommand{\valueFn}[2]{V^{#1}_{#2}}

\newcommand{\QFn}[2]{Q^{#1}_{#2}}

\newcommand{\valuePol}[1]{J\left(#1\right)}

\newcommand{\obs}[0]{o}
\newcommand{\obsSet}[0]{\mathcal{O}}
\newcommand{\obsFnDef}[0]{Z}
\newcommand{\obsFn}[4]{\obsFnDef{}\left(#1, #2, #3, #4\right)}

\newcommand{\belief}[0]{\psi}

\newcommand{\policyBel}[0]{\tilde{\pi}}
\newcommand{\policySetBel}[0]{\tilde{\Pi}}
\newcommand{\valueFnBel}[2]{\tilde{V}^{#1}_{#2}}

\newcommand{\QFnBel}[2]{\tilde{Q}^{#1}_{#2}}

\newcommand{\dataset}[0]{\mathcal{D}}
\newcommand{\policyLEARN}[0]{\hat{\pi}}
\newcommand{\numLearnIter}[0]{N}
\newcommand{\mixfrac}[1]{\beta_{#1}}
\newcommand{\numDatapoints}[0]{m}

\newcommand{\policyOR}[0]{\pi_{\mathrm{OR}}}
\newcommand{\policyMix}[0]{\pi_{\mathrm{mix}}}

\newcommand{\policyMDP}[0]{\pi_{\mathrm{MDP}}}
\newcommand{\policyORBel}[0]{\tilde{\pi}_{\mathrm{OR}}}

\newcommand{\QVal}[0]{Q}

\newcommand{\lossFnPolicyDef}[0]{\mathcal{L}}
\newcommand{\lossFnPolicy}[3]{\lossFnPolicyDef{}\left( #1, #2, #3\right)}

\newcommand{\errclass}[0]{\varepsilon_{\mathrm{class}}}

\newcommand{\errreg}[0]{\varepsilon_{\mathrm{reg}}}

\newcommand{\feature}[0]{f}
\newcommand{\featureIG}[0]{\feature{}_{\mathrm{IG}}}
\newcommand{\featureMot}[0]{\feature{}_{\mathrm{mot}}}

\newcommand{\etal}[0]{et al.}
\newcommand{\aggrevate}[0]{\textsc{AggreVaTe}\xspace}
\newcommand{\algQvalAgg}[0]{\textsc{ExpLOre}\xspace}
\newcommand{\Dagger}[0]{\textsc{DAgger}\xspace}
\newcommand{\RearSideVoxel}[0]{Rear Side Voxel\xspace}
\newcommand{\AverageEntropy}[0]{Average Entropy\xspace}
\newcommand{\OcclusionAware}[0]{Occlusion Aware\xspace}

\title{\LARGE \bf
Learning to Gather Information via Imitation
}

\author{Sanjiban Choudhury$^{1}$, Ashish Kapoor$^{2}$, Gireeja Ranade$^{2}$, Debadeepta Dey$^{2}$
\thanks{This project was formulated and conducted during an internship by Sanjiban Choudhury at Microsoft Research.}
\thanks{$^{1}$Sanjiban Choudhury is with The Robotics Institute, Carnegie Mellon University, USA {\tt\small sanjiban@cmu.edu}}%
\thanks{$^{2}$Ashish Kapoor, Gireeja Ranade and Debadeepta Dey are with Microsoft Research, Redmond, USA {\tt\small \{akapoor, giranade, dedey\}@microsoft.com}}
}

\begin{document}

\maketitle
\thispagestyle{empty}
\pagestyle{empty}

\thispagestyle{empty}
\pagestyle{empty}


\makeatletter
\renewcommand\section{\@startsection{section}{1}{\z@}%
    {1.5ex plus 0.0ex minus 0.5ex}%
    {0.7ex plus 0.0ex minus 0ex}%
    {\normalfont\normalsize\centering\scshape}}%
\renewcommand\subsection{\@startsection{subsection}{2}{\z@}%
    {1.5ex plus 0.0ex minus 0.5ex}%
    {0.7ex plus 0.0ex minus 0ex}%
    {\normalfont\normalsize\itshape}}%
\renewcommand\subsubsection{\@startsection{subsubsection}{3}{\parindent}%
    {0ex plus 0.1ex minus 0.1ex}%
    {0ex}%
    {\normalfont\normalsize\itshape}}%
\renewcommand\paragraph{\@startsection{paragraph}{4}{2\parindent}%
    {0ex plus 0.1ex minus 0.1ex}%
    {0ex}%
    {\normalfont\normalsize\itshape}}%
\makeatother

\setlength \abovedisplayskip{1ex plus0pt minus1pt}
\setlength \belowdisplayskip{1ex plus0pt minus1pt}

\setlength{\skip\footins}{0.5\baselineskip  plus 0.0\baselineskip  minus 0.2\baselineskip} 

\setlength\floatsep{0.9\baselineskip plus0pt minus0.2\baselineskip}                      
\setlength\textfloatsep{0.5\baselineskip plus0pt minus0.4\baselineskip}                
\setlength\abovecaptionskip{0.pt plus0pt minus0pt}                                      

\begin{abstract}
The budgeted information gathering problem - where a robot with a fixed fuel budget is required to maximize the amount of information gathered from the world - appears in practice across a wide range of applications in autonomous exploration and inspection with mobile robots. 
Although there is an extensive amount of prior work investigating effective approximations of the problem, these methods do not address the fact that their performance is heavily dependent on distribution of objects in the world. In this paper, we attempt to address this issue by proposing a novel data-driven imitation learning framework. 

We present an efficient algorithm, \algQvalAgg, that trains a policy on the target distribution to imitate a \emph{clairvoyant oracle} - an oracle that has full information about the world and computes non-myopic solutions to maximize information gathered. We validate the approach on a spectrum of results on a number of 2D and 3D exploration problems that demonstrates the ability of \algQvalAgg to adapt to different object distributions. Additionally, our analysis provides theoretical insight into the behavior of \algQvalAgg. 
Our approach paves the way forward for efficiently applying data-driven methods to the domain of information gathering.
\end{abstract}


\begin{figure}[t!]
    \centering
    \includegraphics[width=\columnwidth]{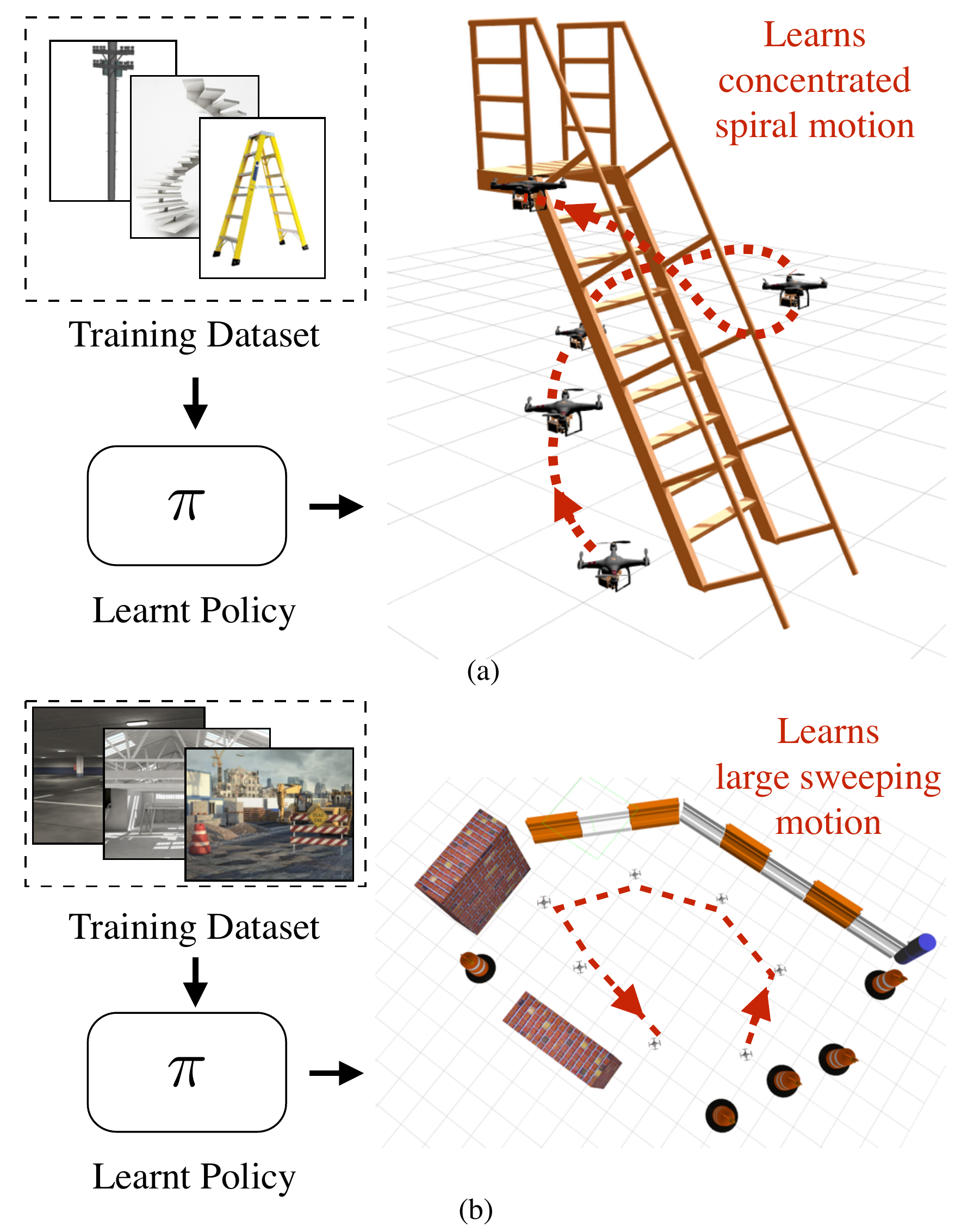}
    \begin{minipage}{\columnwidth}
    \caption{%
    \algQvalAgg trains a policy to gather information. The effectiveness of a policy to gather information depends on the distribution of worlds.
    (a) When the distribution corresponds to a scene containing ladders, the learnt policy executes a helical motion around parts of the ladder already observed as its unlikely that there is information elsewhere.
    (b) When the distribution corresponds to a scene from a construction site, the learnt policy executes a large sweeping motion as information is likely to be dispersed. 
    \label{fig:marquee}
}
    \end{minipage}
    \begin{minipage}{0.01\textwidth}
    \phantomsubcaption{\label{fig:marquee:a}}
    \end{minipage}
    \begin{minipage}{0.01\textwidth}
    \phantomsubcaption{\label{fig:marquee:b}}
    \end{minipage}
\end{figure}

\section{Introduction}
This paper considers the budgeted information gathering problem. Our aim is to maximally explore a world with a robot  that has a budget on the total amount of movement due to battery constraints. This problem fundamentally recurs in mobile robot applications such as autonomous mapping of environments using ground and aerial robots \cite{Charrow-RSS-15,heng2015efficient}, monitoring of water bodies \cite{hollinger2013sampling} 
and inspecting models for 3D reconstruction \cite{isler2016information,hollinger2011active}.

The nature of ``interesting'' objects in an environment and their spatial distribution influence the optimal trajectory a robot might take to explore the environment. As a result, it is important that a robot learns about the type of environment it is exploring as it acquires more information and adapts it's exploration trajectories accordingly. This adaptation must be done online, and we provide such an algorithm in this paper.

To illustrate our point, consider two extreme examples of environments for a particular mapping problem, shown in  Fig.~\ref{fig:marquee}. Consider a robot equipped with a sensor (RGBD camera) that needs to generate a map of an unknown environment. It is given a prior distribution about the geometry of the world, but has no other information. This geometry could include very diverse settings. First it can include a world where there is only one ladder, but the form of the ladder must be explored (Fig.~\ref{fig:marquee:a}), which is a very dense setting. Second, it could include a sparse setting with spatially distributed objects, such as a construction site (Fig.~\ref{fig:marquee:b}). 

The important task for the robot is to now try to infer which type of environment it is in based on the history of measurements, and thus plan an efficient trajectory. At every time step, the robot visits a sensing location and receives a sensor measurement (e.g. depth image) that has some amount of information utility (e.g. surface coverage of objects with point cloud). As opposed to naive lawnmower-coverage patterns, it will be more efficient if the robot could use a policy that maps the history of locations visited and measurements received to decide which location to visit next such that it maximizes the amount of information gathered in the finite amount of battery time available.


The ability of such a learnt policy to gather information efficiently depends on the prior distribution of worlds in which the robot has been shown how to navigate optimally. Fig.~\ref{fig:marquee:a} shows an efficient learnt policy for inspecting a ladder, which executes a helical motion around parts of the ladder already observed to efficiently uncover new parts without searching naively. This is efficient because given the prior distribution the robot learns that information is likely to be geometrically concentrated in a particular volume given it's initial observations of parts of the ladder. Similarly Fig.~\ref{fig:marquee:b} shows an effective policy for exploring construction sites by executing large sweeping motions. Here again the robot learns from prior experience that wide, sweeping motions are efficient since it has learnt that information is likely to be dispersed in such scenarios.

Thus our requirements for an efficient information-gathering policy can be distilled to two points:
\begin{enumerate}
\item \emph{Reasoning about posterior distribution over world maps: } The robot should use the history of movements and measurements to infer a posterior distribution of worlds. This can be used infer locations that are likely to contain information and efficiently plan a trajectory. However the space of world maps is very large, and it is intractable to compute this posterior online. 
\item \emph{Reasoning about non-myopic value of information: } Even if the robot is able to compute the posterior and hence the value of information at a location, it has to be cognizant of the travel cost to get to that location. It needs to exhibit non-myopic behavior to achieve a trade-off that maximizes the overall information gathered. Performing this computationally expensive planning at every step is prohibitively expensive.
\end{enumerate}

Even though its is natural to think of this problem setting as a POMDP, we frame this problem as a novel data-driven imitation learning problem \cite{ross2010efficient}. We propose an algorithm \algQvalAgg (Exploration by Learning to Imitate an Oracle) that trains a policy on a dataset of worlds by imitating a \emph{clairvoyant oracle}. During the training process, the oracle has full information about the world map (and is hence clairvoyant) and plans movements to maximize information. The policy is then trained to imitate these movements as best as it can using partial information from the current history of movements and measurements. As a result of our novel formulation, we are able to sidestep a number of challenging issues in POMDPs like explicitly computing posterior distribution over worlds and planning in belief space. 

Our contributions are as follows
\begin{enumerate}
\item We map the budgeted information gathering problem to a POMDP and present an approach to solve it using imitation learning. 
\item We present an approach to train a policy on the non-stationary distribution of event traces induced by the policy itself. We show that this implicitly results in the policy operating on the posterior distribution of world maps.
\item We show that by imitating an oracle that has access to the world map and thus can plan optimal routes, the policy is able to learn non-myopic behavior. Since the oracle is executed only during train time, the computational burden does not affect online performance.
\end{enumerate}

The remainder of this paper is organized as follows. Section \ref{sec:prob} presents the formal problem, while Section \ref{sec:rel_work} contains relevant work. The algorithm is presented in Section \ref{sec:apprach} and Section \ref{sec:res} presents experimental results. Finally we conclude in Section \ref{sec:conc} with discussions and thoughts on future work.



\section{Problem Statement}
\label{sec:prob}
\subsection{Notation}

Let $\vertexSet$ be a set of nodes corresponding to all sensing locations.
The robot starts at node $\vertexStart$.
Let $\Path = \seq{\vertex}{p}$ be a sequence of nodes (a path) such that $\vertex_1 = \vertexStart$. Let $\PathSet$ be the set of all such paths.
Let $\world \in \worldSet$ be the world map. 
Let $\meas \in \measSet$ be a measurement received by the robot. Let $\measFnDef{}: \vertexSet \times \worldSet \to \measSet$ be a measurement function. When the robot is at node $\vertex$ in a world map $\world$, the measurement $\meas$ received by the robot is $\meas = \measFn{\vertex}{\world}$. 
Let $\utilityFnDef{}: 2^\vertexSet \times \worldSet \to \real_{\geq 0}$ be a utility function. For a path $\Path$ and a world map $\world$, $\utilityFn{\Path}{\world}$ assigns a utility to executing the path on the world. Note that $\utilityFnDef{}$ is a set function.
Given a node $\vertex \in \vertexSet$, a set of nodes $\vertexGroup \subseteq \vertexSet$ and world $\world$, the discrete derivative of the utility function $\utilityFnDef$ is $\marginalGain{\vertex \mid \vertexGroup}{\world} = \utilityFn{\vertexGroup \cup \{ \vertex \}}{\world} - \utilityFn{\vertexGroup}{\world}$
Let $\costFnDef{}: \PathSet \times \worldSet \to \real_{\geq 0}$ be a travel cost function. For a path $\Path$ and a world map $\world$, $\costFn{\Path}{\world}$ assigns a travel cost to executing the path on the world. 

\subsection{Problem Formulation}

We first define the problem setting when the world map is fully known.

\begin{problem}[Fully Observable World Map; Constrained Travel Cost] \label{prob:fully_obs:cons}
Given a world map $\world$, a travel cost budget $\costBudget$ and a time horizon $T$, find a path $\Path$ that maximizes utility subject to travel cost and cardinality constraints.
\begin{equation}
\begin{aligned}
\argmaxprob{\Path \in \PathSet} \quad & \utilityFn{\Path}{\world} \\
\suchthat                             & \costFn{\Path}{\world} \leq \costBudget \\
                                      &  \card{\Path} \leq T+1
\end{aligned}
\end{equation}
\end{problem}

Now, consider the setting where the world map $\world$ is unknown. Given a prior distribution $P(\world)$, it can be inferred only via the measurements $\meas_i$ received as the robot visits nodes $\vertex_i$. Hence, instead of solving for a fixed path, we compute a policy that maps history of measurements received and nodes visited to decide which node to visit. 

\begin{problem}[Partially Observable World Map; Constrained Travel Cost] \label{prob:partially_obs:cons}
Given a distribution of world maps, $P(\world)$, a travel cost budget $\costBudget$ and a time horizon $T$, find a policy that at time $t$, maps the history of nodes visited $\{ \vertex_i \}_{i=1}^{t-1}$ and measurements received $\{ \meas_i \}_{i=1}^{t-1}$ to compute node $\vertex_t$ to visit at time $t$, such that the expected utility is maximized subject to travel cost and cardinality constraints. 
\end{problem}

\subsection{Mapping to MDP and POMDP}

\subsubsection{Mapping fully observable problems to MDP}

The Markov Decision Process (MDP) is a tuple $(\stateSet, \worldSet, \actionSet, \transFnDef, \rewardFnDef, T)$ defined upto a fixed finite horizon $T$. It is defined over an augmented state space comprising of the ego-motion state space $\stateSet$ (which we will refer to as simply  the state space) and the space of world maps $\worldSet$.

Let $\state_t \in \stateSet$ be the state of the robot at time $t$. It is defined as the set of nodes visited by the robot upto time $t$, $\state_t = \seq{\vertex}{t}$. This implies the dimension of the state space is exponentially large in the space of nodes, $\stateSet = 2 ^{\card{\vertexSet}}$. The initial state $\state_1 = \vertexStart$ is the start node.
Let $\action_t \in \actionSet$ be the action executed by the robot at time $t$. It is defined as the node visited at time $t+1$, $\action_t = \vertex_{t+1}$. The set of all actions is defined as $\actionSet = \vertexSet$. Given a world map $\world$, when the robot is at state $\state$ the utility of executing an action $\action$ is $\utilityFn{\state \cup \action}{\world}$.
Let $\actionSetFeas{\state}{\world} \subset \actionSet$ be the set of feasible actions that the robot can execute when in state $\state$ in a world map $\world$. This is defined as follows
\begin{equation}
   \actionSetFeas{\state}{\world} = \setst{\action}{\action \in \actionSet, \costFn{\state \cup \action}{\world} \leq \costBudget}
\end{equation} 
Let $\transFn{\state}{\action}{\state'} = P(\state' | \state, \action)$ be the state transition function. In our setting, this is the deterministic function $\state' = \state \cup \action$. 
Let $\rewardFn{\state}{\world}{\action} \in [0,1]$ be the one step reward function. It is defined as the normalized marginal gain of the utility function, $\rewardFn{\state}{\world}{\action} = \frac{\marginalGain{\action \mid \state}{\world}}{\utilityFn{\actionSet}{\world}}$.
Let $\policy(\state, \world) \in \policySet$ be a policy that maps state $\state$ and world map $\world$ to a feasible action $\action \in \actionSetFeas{\state}{\world}$. 
The value of executing a policy $\policy$ for $t$ time steps on a world $\world$ starting at state $\state$.
\begin{equation}
\valueFn{\policy}{t}(\state, \world) = \sum\limits_{i=1}^{t} \expect{\state_i \sim P(\state' \mid \state, \policy, i) }{\rewardFn{\state_i}{\world}{\policy(\state_i,\world)}}
\end{equation}
where $P(\state' | \state, \policy, i)$ is the distribution of states at time $i$ starting from $\state$ and following policy $\policy$.
The state action value $\QFn{\policy}{t}(\state, \world, \action)$ is the value of executing action $\action$ in state $\state$ in world $\world$ and then following 
the policy $\policy$ for $t-1$ timesteps

\begin{equation}
\QFn{\policy}{t}(\state, \world, \action) = \rewardFn{\state}{\world}{\action} + \expect{\state' \sim P(\state' \mid \state, \action)}{\valueFn{\policy}{t-1}(\state', \world)} 
\end{equation}

The value of a policy $\policy$ for $T$ steps on a distribution of worlds $P(\world)$ and starting states $P(\state)$

\begin{equation}
\valuePol{\policy} = \expect{\state \sim P(\state), \world \sim P(\world)}{\valueFn{\policy}{T}(\state, \world)}
\end{equation}

The optimal MDP policy is $\policyMDP = \argmax\limits_{\policy \in \policySet}\valuePol{\policy}$.

\subsubsection{Mapping partially observable problems to POMDP}

The Partially Observable Markov Decision Process (POMDP) is a tuple $(\stateSet, \worldSet, \actionSet, \transFnDef, \rewardFnDef, \obsSet, \obsFnDef, T)$. The first component of the augmented state space, the ego motion state space $\stateSet$, is fully observable. The second component, the space of world maps $\worldSet$, is partially observable through observations received.

Let $\obs_t \in \obsSet$ be the observation at time step $t$. This is defined as the measurement received by the robot $\obs_t = \meas_t$. Let $\obsFn{\state}{\action}{\world}{\obs} = P(\obs | \state, \action, \world)$ be the probability of receiving an observation $\obs$ given the robot is at state $\state$ and executes action $\action$. In our setting, this is the deterministic function $\obs = \measFn{\state \cup \action}{\world}$.

Let the belief at time $t$, $\belief_t$, be the history of state, action, observation tuple received so far, i.e. $\{(\state_i, \action_i, \obs_i)\}_{i=1}^{t}$. Note that this differs from the conventional use of the word belief which would usually imply a distribution. However, we use belief here to refer to the history of state, action, observations conditioned on which one can infer the posterior distribution of world maps $P(\phi)$. 
Let the belief transition function be $P(\belief' | \belief, \action)$.
Let $\policyBel(\state, \belief) \in \policySetBel$ be a policy that maps state $\state$ and belief $\belief$ to a feasible action $\action \in \actionSetFeas{\state}{\world}$. 
The value of executing a policy $\policyBel$ for $t$ time steps starting at state $\state$ and belief $\belief$ is
\begin{equation}
\valueFnBel{\policyBel}{t}(\state, \belief) = \sum\limits_{i=1}^{t} \expect{
\substack{\belief_i \sim P(\belief' \mid \belief, \policyBel, i), \\
\world \sim P(\world \mid \belief_i) \\ \state_i \sim P(\state' \mid \state, \policyBel, i)}
 }{\rewardFn{\state_i}{\world}{\policyBel(\state_i,\belief_i)}}
\end{equation}
where $P(\belief' | \belief, \policyBel, i)$ is the distribution of beliefs at time $i$ starting from $\belief$ and following policy $\policyBel$. 
$P(\world | \belief_i)$ is the posterior distribution on worlds given the belief $\belief_i$. 
Similarly the action value function $\QFnBel{\policyBel}{t}$ is defined as 
\begin{equation}
\begin{aligned}
\label{eq:pomdp:q}
\QFnBel{\policyBel}{t}(\state, \belief, \action) =& \expect{\world \sim P(\world \mid \belief)}{\rewardFn{\state}{\world}{\action}} + \\
                        &\expect{\belief' \sim P(\belief' \mid \belief, \action), \state' \sim P(\state' \mid \state, \action)}{\valueFnBel{\policyBel}{t-1}(\state', \belief')}
\end{aligned}
\end{equation}

The optimal POMDP policy can be expressed as 
\begin{equation}
\label{eq:pomdp:opt_policy}
\policy^* = \argmax\limits_{\policyBel \in \policySetBel} 
\expect{
\substack{t\sim U(1:T), \\
 \state \sim P(\state \mid \policyBel, t), \\
 \belief \sim P(\belief \mid \policyBel, t)}
 }{\QFnBel{\policyBel}{T-t+1}(\state, \belief, \policyBel(\state,\belief))}
\end{equation}
where $U(1:T)$ is a uniform distribution over the discrete interval $\{1, 2, \dots, T\}$, $P(\state \mid \policyBel, t)$ is the distribution of states following policy $\policyBel$ for $t$ steps, 
$P(\belief \mid \policyBel, t)$ is the distribution of belief following policy $\policyBel$ for $t$ steps.
The value of a policy $\policyBel \in \policySetBel$ for $T$ steps on a distribution of worlds $P(\world)$, starting states $P(\state)$ and starting belief $P(\belief)$
\begin{equation}
\valuePol{\policyBel} = \expect{\state \sim P(\state), \belief \sim P(\belief)}{\valueFn{\policyBel}{T}(\state, \belief)}
\end{equation}
where the posterior world distribution $P(\world \mid \belief)$ uses $P(\world)$ as prior. 


\section{Related Work}
\label{sec:rel_work}

\begin{figure*}[!htp]
    \centering
    \includegraphics[width=\textwidth]{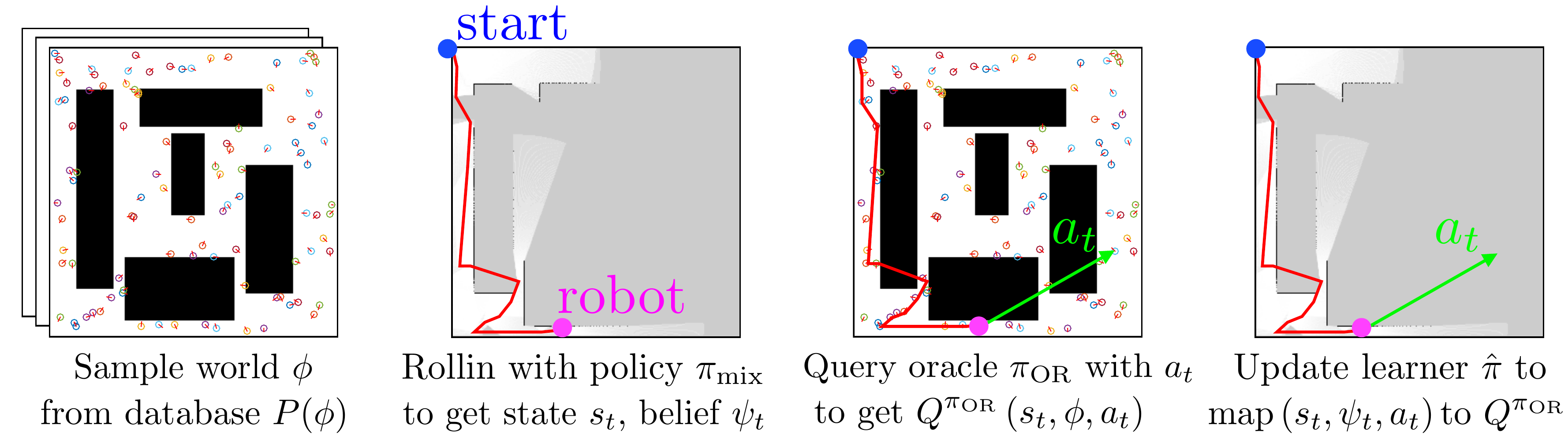}
    \caption{%
        Overview of \algQvalAgg. The algorithm iteratively trains a learner $\policyLEARN$ to imitate a clairvoyant oracle $\policyOR$. A world map $\world$ is sampled from database representing $P(\world)$. A mixture policy $\policyMix$ of $\policyLEARN$ and $\policyOR$ is used to roll-in on $\world$ for a random timestep $t$ to get state $\state_t$ and belief $\belief_t$. A exploratory action $a_t$ is chosen.
        The clairvoyant oracle $\policyOR$ is given full access to world map $\world$ to compute the cumulative reward to go $\QFn{\policyOR}{T-t+1}\left(\state_t, \world, \action_t \right)$. This datapoint comprising of the belief from roll in $\psi_t$, the state $\state_t$, the action $\action_t$ and the value $\QVal^{\policyOR}$ is used to create a cost sensitive classification problem that updates the learner $\policyLEARN$.
        \fullFigGap}
    \label{fig:alg:overview}
\end{figure*}%

Problem~\ref{prob:fully_obs:cons} is a submodular function optimization (due to nature of $\utilityFnDef$) subject to routing constraints (due to $\costFnDef$). In absence of this constraint, there is a large body of work on near optimality of greedy strategies by Krause \etal \cite{krause2012submodular,krause2008efficient,Krause:2007:NOS:1619797.1619913} - however naive greedy approaches can perform arbitrarily poorly. Chekuri and Pal \cite{chekuri2005recursive} propose a quasi-polynomial time recursive greedy approach to solving this problem. Singh \etal \cite{singh2007efficient} show how to scale up the approach to multiple robots. However, these methods are slow in practice. Iyer and Bilmes \cite{iyer2013submodular} solve a related problem of submodular maximization subject to submodular knapsack constraints using an iterative greedy approach. This inspires Zhang and Vorobeychik \cite{zhang2016submodular} to propose an elegant generalization of the cost benefit algorithm (GCB) which we use as an oracle in this paper. Yu \etal \cite{yu2014correlated} frame the problem as a correlated orienteering problem and propose a mixed integer based approach - however only correlations between neighboring nodes are considered. Hollinger and Sukhatme \cite{hollinger2013sampling} use sampling based approaches which require a lot of evaluations of the utility function in practice.

Problem~\ref{prob:partially_obs:cons} in the absence of the travel cost constraint can be efficiently solved using the framework of adaptive submodularity developed by Golovin \etal \cite{golovin2011adaptive,golovin2010near} as shown by Javdani \etal \cite{Javdani_2013_7419,Javdani_2014_7555} and Chen \etal \cite{AAAI159841,DBLP:journals/corr/ChenHK16a}. Hollinger \etal \cite{hollinger2012active,hollinger2011active} propose a heuristic based approach to select a subset of informative nodes and perform minimum cost tours. Singh \etal \cite{Singh:2009:NAI:1661445.1661741} replan every step using a non-adaptive information path planning algorithm. Such methods suffer when the adaptivity gap is large \cite{hollinger2012active}. Inspired by adaptive TSP approaches by Gupta \etal \cite{gupta2010approximation}, Lim \etal  \cite{lim2016adaptive,NIPS2015_6005} propose recursive coverage algorithms to learn policy trees. However such methods cannot scale well to large state and observation spaces. Heng \etal \cite{heng2015efficient} make a modular approximation of the objective function. Isler \etal \cite{isler2016information} survey a broad number of myopic information gain based heuristics that work well in practice but have no formal guarantees.

Online POMDP planning also has a large body of work (\cite{kaelbling1998planning,shani2013survey,kurniawati2008sarsop}. Although there exists fast solvers such as POMCP (Silver and Veness \cite{silver2010monte}) and DESPOT (Somani \etal \cite{somani2013despot}), the space of world maps is too large for online planning.




\section{Approach}
\label{sec:apprach}

\subsection{Overview}
Fig.~\ref{fig:alg:overview} shows an overview of our approach. The central idea is as follows - we train a policy to imitate an algorithm that has access to the world map at train time. 
The policy $\policyLEARN(\state, \belief)$ maps features extracted from state $\state$ and belief $\belief$ to an action $\action$.
The algorithm that is being imitated has access to the corresponding world map $\world$. 

\subsection{Imitation Learning}

We now formally define imitation learning as applied to our setting. Let $\policyBel \in \policySetBel$ be a policy defined on a pair of state and belief $(\state, \belief)$. Let \emph{roll-in} be the process of executing a policy $\policyBel$ from the start upto a certain time horizon. Similarly \emph{roll-out} is the process of executing a policy from the current state and belief till the end. Let $P(\state | \policyBel)$ be the distribution of states induced by roll-in with policy $\policyBel$. Let $P(\belief | \policyBel)$ be the distribution of belief induced by roll-in with policy $\policyBel$.

Let $\lossFnPolicy{\state}{\belief}{\policyBel}$ be the loss of a policy $\policyBel$ when executed on state $\state$ and belief $\policyBel$. This loss function implicitly captures how well policy $\policyBel$ imitates a reference policy (such an an oracle algorithm).
Our goal is to find a policy $\policyLEARN$ which minimizes the observed loss under its own induced distribution of state and beliefs.

\begin{equation}
\label{eq:imitation_learning}
\policyLEARN = \argmin\limits_{\policyBel \in \policySetBel} \expect{ 
\state \sim P(\state \mid \policyBel), 
\belief \sim P(\belief \mid \policyBel)}
{\lossFnPolicy{\state}{\belief}{\policyBel}}
\end{equation} 

This is a non-i.i.d supervised learning problem. Ross and Bagnell \cite{ross2010efficient} show how such problems can be reduced to no-regret online learning using dataset aggregation (\Dagger). The loss function they consider $\lossFnPolicyDef$ is a mis-classification loss with respect to what the expert demonstrated. Ross and Bagnell \cite{ross2014reinforcement} extend the approach to the reinforcement learning setting where $\lossFnPolicyDef$ is the cost-to-go of an oracle reference policy by aggregating values to imitate (\aggrevate).

\subsection{Solving POMDP via Imitation of a Clairvoyant Oracle}

When (\ref{eq:pomdp:opt_policy}) is compared to the imitation learning framework in (\ref{eq:imitation_learning}), we see that in addition to the induced state belief distributions, the loss function analogue $\lossFnPolicy{\state}{\belief}{\policyBel}$  is $\QFnBel{\policyBel}{T-t+1}(\state, \belief, \policyBel(\state,\belief))$. This implies rolling out with policy $\policyBel$. For poor policies $\policyBel$, the action value estimate $\QFnBel{\policyBel}{T-t+1}$ would be very different from optimal values $\QFnBel{\policy^*}{T-t+1}$. 

In our approach, we alleviate this problem by defining a surrogate value functions to imitate - the cumulative reward gathered by a clairvoyant oracle. 
\begin{definition}[Clairvoyant Oracle]
Given a distribution of world map $P(\world)$, a clairvoyant oracle $\policyOR(\state, \world)$ is a policy that maps state $\state$ and world map $\world$ to a feasible action $\action \in \actionSetFeas{\state}{\world}$ such that it approximates the optimal MDP policy, $\policyOR \approx \policyMDP = \argmaxprob{\policy \in \policySet}\valuePol{\policy}$.
\end{definition}

The term \emph{clairvoyant} is used because the oracle has full access to the world map $\world$ at train time. The oracle can be used to compute state action value as follows

\begin{equation}
\QFn{\policyOR}{t}(\state, \world, \action) = \rewardFn{\state}{\world}{\action} + \expect{\state' \sim P(\state' \mid \state, \action)}{\valueFn{\policyOR}{t-1}(\state', \world)} 
\end{equation}

Our approach is to imitate the oracle during training. This implies that we train a policy $\policyLEARN$ by solving the following optimization problem

\begin{equation}
\label{eq:imitateClairvoyantOracle}
\policyLEARN = \argmax\limits_{\policyBel \in \policySetBel} \expect{
\substack{t\sim U(1:T), \\
\state \sim P(\state \mid \policyBel, t), \\
\world \sim P(\world), \\
\belief \sim P(\belief \mid \world, \policyBel, t)}}
{\QFn{\policyOR}{T-t+1}(\state, \world, \policyBel(\state,\belief))}
\end{equation}

\subsection{Algorithm}

\begin{algorithm}
\caption{\algQvalAgg : Imitation Learning of Oracle \label{alg:qvalAgg}}
\begin{algorithmic}[1]
\State Initialize $\dataset \gets \emptyset$, $\policyLEARN_1$ to any policy in $\policySetBel$ \label{alg:qvalAgg:init}
\For{$i=1$ \textbf{to} $\numLearnIter$}
\State Initialize sub dataset $\dataset_i \gets \emptyset$\; \label{alg:qvalAgg:initSub}
\State Let roll in policy be $\policyMix = \mixfrac{i} \policyOR + (1-\mixfrac{i}) \policyLEARN_i$ \label{alg:qvalAgg:mixPol}
\State Collect $m$ data points as follows:
\For{$j=1$ \textbf{to} $\numDatapoints$}
\State Sample world map $\world$ from dataset $P(\world)$ \label{alg:qvalAgg:sampleWorld}
\State Sample uniformly $t \in \{1,2,\dots,T\}$ \label{alg:qvalAgg:sampleTime}
\State Assign initial state $\state_1 = \vertexStart$ \label{alg:qvalAgg:initialState}
\State Execute $\policyMix$ up to time $t-1$ to reach $\pair{\state_t}{\belief_t}$ \label{alg:qvalAgg:rollin}
\State Execute any action $\action_t \in \actionSetFeas{\state_t}{\world}$ \label{alg:qvalAgg:takeAction}
\State Execute oracle $\policyOR$ from $t+1$ to $T$ on $\world$ \label{alg:qvalAgg:execOracle}
\State Collect value to go $\QVal_i^{\policyOR} =  {\QFn{\policyOR}{T-t+1}(\state_t, \world, \action_t)}$ \label{alg:qvalAgg:collectVal}
\State $\dataset_i \gets \dataset_i \cup \{\state_t, \belief_t, \action_t, t, \QVal_i^{\policyOR}\}$ \label{alg:qvalAgg:aggrSubData}
\EndFor
\State Aggregate datasets: $\dataset \gets \dataset \bigcup \dataset_i$ \label{alg:qvalAgg:aggrData}
\State Train cost-sensitive classifier $\policyLEARN_{i+1}$ on $\dataset$\\ \label{alg:qvalAgg:updateLearner}
~~~~ (\emph{Alternately: use any online learner $\policyLEARN_{i+1}$ on $\dataset_i$})
\EndFor
\State \textbf{Return} best $\policyLEARN_i$ on validation
\end{algorithmic}
\end{algorithm}

Alg.~\ref{alg:qvalAgg} describes the \algQvalAgg algorithm. The algorithm iteratively trains a sequence of learnt policies $\seq{\policyLEARN}{\numLearnIter}$ by aggregating data for an online cost-sensitive classification problem. 

$\policyLEARN_1$ is initialized as a random policy (Line~\ref{alg:qvalAgg:init}). At iteration $i$, the policy that is used to roll-in is a mixture policy of learnt policy $\policyLEARN_i$ and the oracle policy $\policyOR$ (Line~\ref{alg:qvalAgg:mixPol}) using mixture parameter $\mixfrac{i}$. A set of $\numDatapoints$ cost-sensitive classification datapoints are captured as follows: a world $\world$ is sampled (Line~\ref{alg:qvalAgg:sampleWorld}). The $\policyMix$ is used to roll-in upto a random time from an initial state to reach $\pair{\state_t}{\belief_t}$ (Lines~\ref{alg:qvalAgg:sampleTime}--\ref{alg:qvalAgg:rollin}). An exploratory action is selected (Line~\ref{alg:qvalAgg:takeAction}). The clairvoyant oracle is given full access to $\world$ and asked to roll-out and provide an action value $\QVal^{\policyOR}$ (Lines~\ref{alg:qvalAgg:execOracle}--\ref{alg:qvalAgg:collectVal}).
$\{\state_t, \belief_t, \action_t, t, \QVal_i^{\policyOR}\}$ is added to a dataset $\dataset_i$ of cost-sensitive classification problem and the process is repeated (Line~\ref{alg:qvalAgg:aggrSubData}). $\dataset_i$ is appended to the original dataset $\dataset$ and used to train an updated learner $\policyLEARN_{i+1}$ (Lines~\ref{alg:qvalAgg:aggrData}--\ref{alg:qvalAgg:updateLearner}). The algorithm returns the best learner from the sequence based on performance on a held out validation set (or alternatively returns a mixture policy of all learnt policies).
One can also try variants where all actions $\action \in \actionSetFeas{\state_t}{\world}$ are executed, or an online learner is used to update $\policyLEARN$ instead of dataset aggregation \cite{ross2014reinforcement}.


\section {Analysis}
\label{sec:analysis}

Following the analysis style of \aggrevate \cite{ross2014reinforcement}, we first introduce a \emph{hallucinating oracle}.
\begin{definition}[Hallucinating Oracle]
Given a prior distribution of world map $P(\world)$, a hallucinating oracle $\policyORBel$ computes the instantaneous posterior distribution over world maps and takes the action with the highest expected value. 
\begin{equation}
  \policyORBel = \argmax\limits_{\action \in \actionSet} \expect{\world \sim P(\world \mid \belief)}{\QFn{\policyOR}{t}(\state, \world, \action)}
\end{equation}
\end{definition}
While the hallucinating oracle is not the optimal POMDP policy (\ref{eq:pomdp:opt_policy}), it is an effective policy for information gathering as alluded to in \cite{Koval-RSS-14} and we now show that we effectively imitate it.
\begin{lemma}
The policy optimization rule in (\ref{eq:imitateClairvoyantOracle}) is equivalent to 
\begin{equation*}
\policyLEARN = \argmax\limits_{\policyBel \in \policySetBel} \expect{
\substack{t\sim U(1:T), \\
\state \sim P(\state \mid \policyBel, t), \\
\world \sim P(\world), \\
\belief \sim P(\belief \mid \world, \policyBel, t)}}
{\QFn{\policyORBel}{T-t+1}(\state, \world, \policyBel(\state,\belief))}
\end{equation*}
by using the fact that 
$\expect
{\world \sim P(\world)}
{\QFn{\policyOR}{t}(\state, \world, \action)} = 
\expect{
\substack{\world \sim P(\world),\\
\belief \sim P(\belief \mid \world, \policyBel, t)}}
{\QFn{\policyORBel}{t}(\state, \world, \action)}
$
\end{lemma}
Consequently our learnt policy has the following guarantee
\begin{theorem}
N iterations of \algQvalAgg, collecting $m$ regression examples per iteration guarantees that with probability at least $1-\delta$
\begin{equation*}
\begin{aligned}
  \valuePol{\policyLEARN} \geq & \valuePol{\policyORBel} \\
  & - 2\sqrt{\abs{\actionSet}}T\sqrt{\errclass + \errreg 
  + O \left(\sqrt{\log \left(\nicefrac{ \left( \nicefrac{1}{\delta} \right)}{Nm}\right)} \right)} \\
  & - O \left(\frac{T^2 \log T}{\alpha N}\right)
\end{aligned}
\end{equation*}
where $\errreg$ is the empirical average online learning regret on the training regression examples collected over the iterations and 
$\errclass$ is the empirical regression regret of the best regressor in the policy class.
\end{theorem}
For both proofs, refer to \cite{ExploreProof}.


\section{Experimental Results}
\label{sec:res}

\begin{figure*}[t]
    \centering
    \includegraphics[page=1,width=\textwidth]{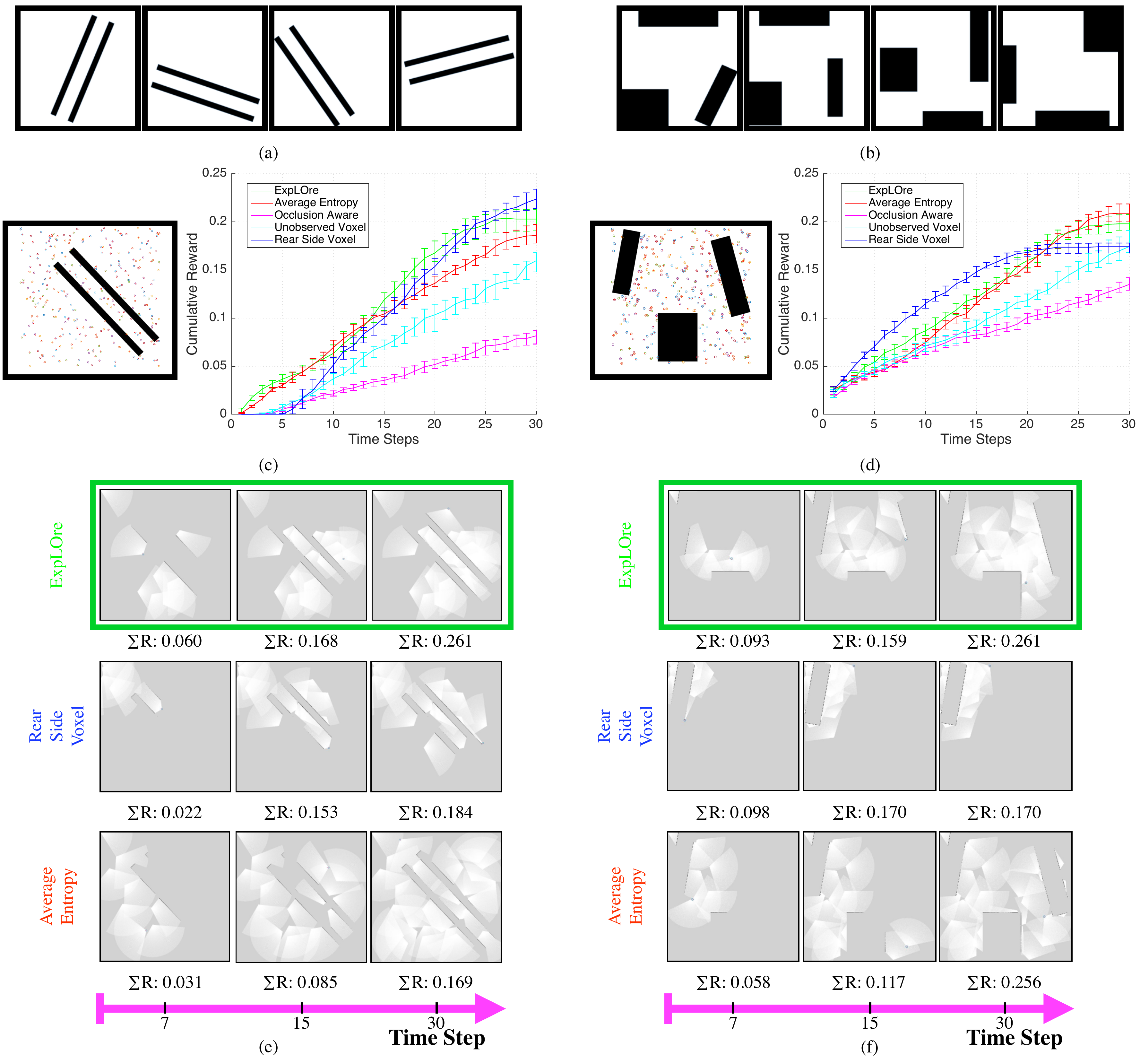}
\begin{minipage}{\textwidth}
    \caption{%
    Comparison of \algQvalAgg with baseline heuristics on a 2D exploration problem on 2 different datasets - dataset 1 (concentrated information) and dataset 2 (distributed information). The problem details are: $T=30, \costBudget=2500, |\actionSet|=300$.
    Sample world maps from (a) dataset 1 and (b) dataset 2. Training dataset is created with $10$ world maps, each with $10$ random node sets to create a dataset size of $100$. 
    Test results on $1$ representative world map with $100$ random node sets are shown for (c) dataset 1 and (d) dataset 2.
    A sample test instance is shown along with a plot of cumulative reward with time steps for \algQvalAgg and other baseline heuristics. The error bars show $95\%$ confidence intervals. 
    Snapshots of execution of \algQvalAgg, \RearSideVoxel and \AverageEntropy are shown for (e) dataset 1 and (f) dataset 2. The snapshots show the evidence grid at time steps $7, 15$ and $30$. 
        \label{fig:results:matlab}
        \fullFigGap}
\end{minipage}

            \begin{minipage}{0.01\textwidth}
\phantomsubcaption{\label{fig:results:matlab:a}}
\end{minipage}
\begin{minipage}{0.01\textwidth}
\phantomsubcaption{\label{fig:results:matlab:b}}
\end{minipage}
    \begin{minipage}{0.01\textwidth}
\phantomsubcaption{\label{fig:results:matlab:c}}
\end{minipage}
\begin{minipage}{0.01\textwidth}
\phantomsubcaption{\label{fig:results:matlab:d}}
\end{minipage}
\begin{minipage}{0.01\textwidth}
\phantomsubcaption{\label{fig:results:matlab:e}}
\end{minipage}
\begin{minipage}{0.01\textwidth}
\phantomsubcaption{\label{fig:results:matlab:f}}
\end{minipage}
\end{figure*}%

\begin{figure*}[t]
    \centering
    \includegraphics[page=1,width=\textwidth]{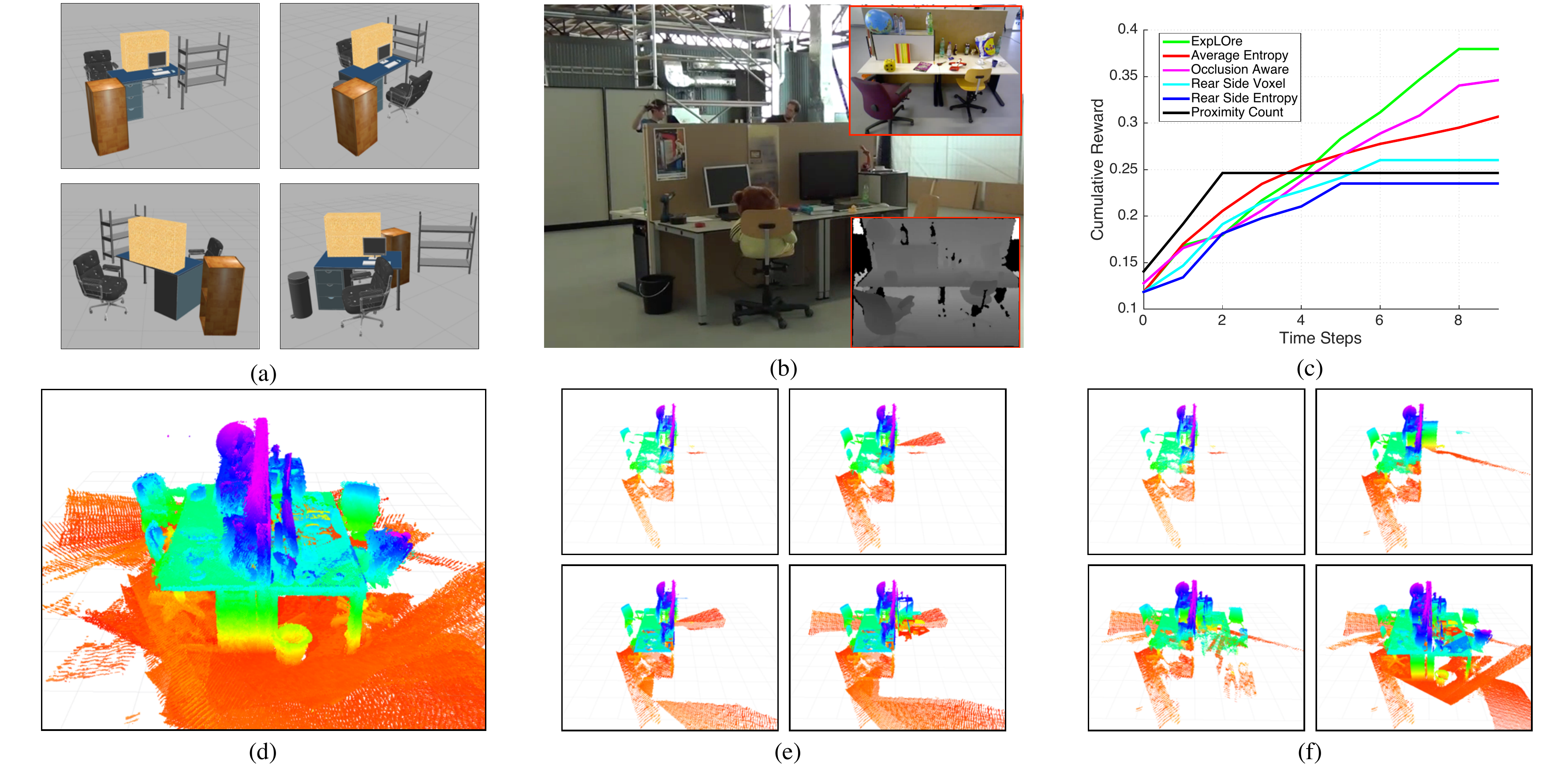}

\begin{minipage}{\textwidth}
    \caption{%
    Comparison of \algQvalAgg with baseline heuristics on a 3D exploration problem where training is done on simulated world maps and testing is done on a real dataset of an office workspace. The problem details are: $T=10$, $\costBudget=12$, $|\actionSet|=50$. 
    (a) Samples from $100$ simulated worlds resembling an office workspace created in Gazebo.
    (b) Real dataset collected by \cite{sturm12iros} using a RGBD camera. 
    (c) Plot of cumulative reward with time steps for \algQvalAgg and baseline heuristics on the real dataset.
    (d) The 3D model of the real office workspace formed by cumulating measurements from all poses. 
    (e) Snapshots of execution of \OcclusionAware heuristic at time steps $1,3, 5, 9$.
    (f) Snapshots of execution of \algQvalAgg heuristic at time steps $1,3, 5, 9$. 
    \label{fig:results:cpp}
        \fullFigGap}
\end{minipage}

                \begin{minipage}{0.01\textwidth}
\phantomsubcaption{\label{fig:results:cpp:a}}
\end{minipage}
\begin{minipage}{0.01\textwidth}
\phantomsubcaption{\label{fig:results:cpp:b}}
\end{minipage}
    \begin{minipage}{0.01\textwidth}
\phantomsubcaption{\label{fig:results:cpp:c}}
\end{minipage}
\begin{minipage}{0.01\textwidth}
\phantomsubcaption{\label{fig:results:cpp:d}}
\end{minipage}
\begin{minipage}{0.01\textwidth}
\phantomsubcaption{\label{fig:results:cpp:e}}
\end{minipage}
\begin{minipage}{0.01\textwidth}
\phantomsubcaption{\label{fig:results:cpp:f}}
\end{minipage}
\end{figure*}%

\begin{figure*}[t]
    \centering
    \includegraphics[page=1,width=\textwidth]{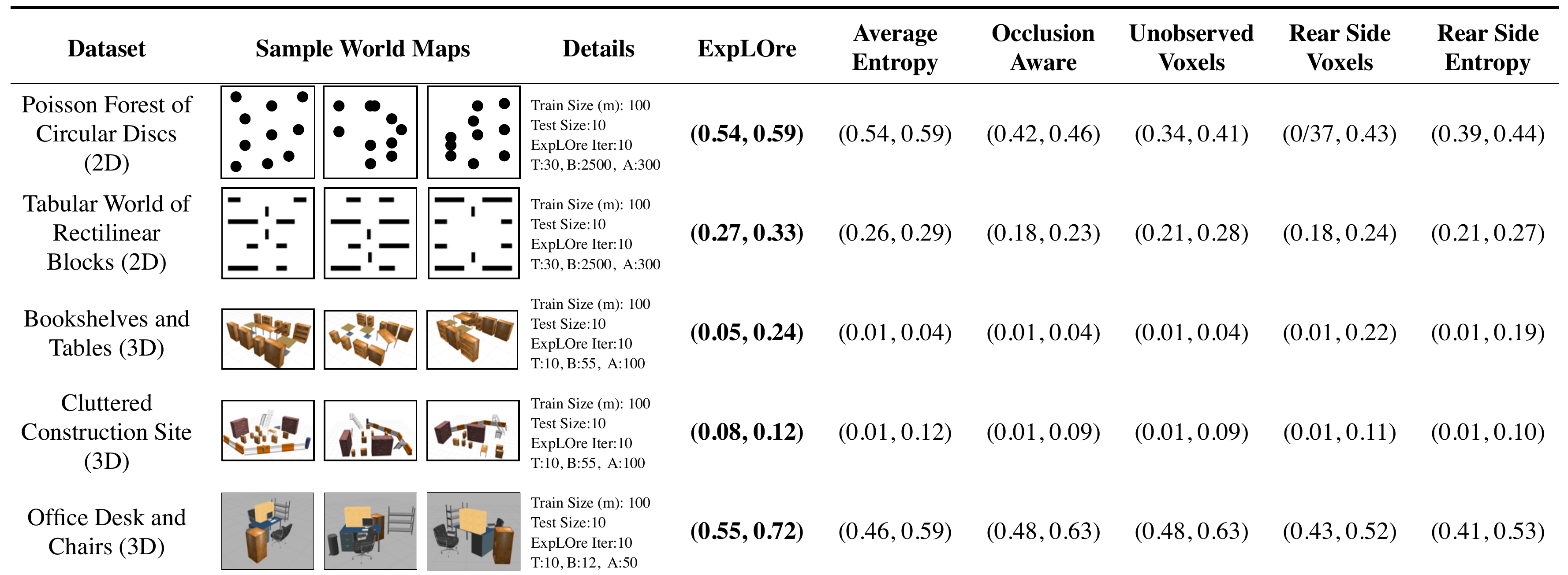}
    \caption{%
    Comparison of \algQvalAgg with baseline heuristics on a number of experiments both 2D and 3D. Each row corresponds to different datasets. The columns contain information about the dataset, representative pictures and performance results for all algorithms. The numbers are the lower and upper confidence (for 95\% CI) of cumulative reward at the final time step. The algorithm with the highest median performance is emphasized in bold for each dataset. 
          \fullFigGap}
    \label{fig:results:extra}
\end{figure*}%

\subsection{Implementation Details}
Our implementation is open source and available for MATLAB and C++ \url{goo.gl/HXNQwS}. 

\subsubsection{Problem Details} 
The utility function $\utilityFnDef$ is selected to be a fractional coverage function (similar to \cite{isler2016information}) which is defined as follows.
The world map $\world$ is represented as a voxel grid representing the surface of a 3D model. The sensor measurement $\measFn{\vertex}{\world}$ at node $\vertex$ is obtained by raycasting on this 3D model. A voxel of the model is said to be `covered' by a measurement received at a node if a point lies in that voxel. The coverage of a path $\Path$ is the fraction of covered voxels by the union of measurements received when visiting each node of the path. The travel cost function $\costFnDef$ is chosen to be the euclidean distance.
The values of total time step $T$ and travel budget $\costBudget$ varies with problem instances and are specified along with the results.
\subsubsection{Oracle Algorithm}
We use the generalized cost benefit algorithm (GCB) \cite{zhang2016submodular} as the oracle algorithm owing to its small run times and acceptable solution qualities.
\subsubsection{Learning Details}

The tuple $\left(\state, \action, \belief \right)$ is mapped to a vector of features $\feature =  \bbm \featureIG^T & \featureMot^T \ebm^T$. The feature vector $\featureIG$ is a vector of information gain metrics as described in \cite{isler2016information}. $\featureMot$ encode the relative rotation and translation required to visit a node. We use random forest \cite{liaw2002classification} to regress to $\QVal$ values from features $\feature$. The learning details are specified in Table.~\ref{tab:learning_details}.

\begin{table}[!htbp]
\centering
\caption{Learning Details}
\begin{tabulary}{\textwidth}{LCCCC}\toprule
    {\bf Problem}   & {\bf Train}         & {\bf Test}        & {\bf \algQvalAgg}     & {\bf Feature} \\
                    & {\bf Dataset $m$}   & {\bf Dataset}     & {\bf Iterations $N$}  & {\bf Dimension $|\feature|$} \\ \midrule
    2D  & $100$  & $100$ & $100$  & $16$      \\
    3D  & $100$  & $10$ & $10$  & $16$      \\ \bottomrule
\end{tabulary}
\label{tab:learning_details}
\end{table}

\subsubsection{Baseline}

For baseline policies, we compare to the class of information gain heuristics discussed in \cite{isler2016information}. The heuristics are remarkably effective, however, their performance depends on the distribution of objections in a world map. As \algQvalAgg uses these heuristic values as part of its feature vector, it will implicitly learn a data driven trade-off between them. 

\subsection{2D Exploration}

We create a set of 2D exploration problems to gain a better understanding of the behavior of the \algQvalAgg and baseline heuristics. A dataset comprises of 2D binary world maps, uniformly distributed nodes and a simulated laser. The training size is $100$, $T=30$, $\costBudget = 2500$.

\subsubsection{Dataset 1: Concentrated Information}

Fig.~\ref{fig:results:matlab:a} shows a dataset created by applying random affine transformations to a pair of parallel lines. 
This dataset is representative of information being concentrated in a particular fashion.
Fig.~\ref{fig:results:matlab:c} shows a comparison of \algQvalAgg with baseline heuristics. The heuristic \RearSideVoxel performs the best, while \algQvalAgg is able to match the heuristic. 
Fig.~\ref{fig:results:matlab:e} shows progress of \algQvalAgg along with two relevant heuristics - \RearSideVoxel and \AverageEntropy. \RearSideVoxel takes small steps focusing on exploiting viewpoints along the already observed area. \AverageEntropy aggressively visits unexplored area which is mainly free space. \algQvalAgg initially explores the world but on seeing parts of the lines reverts to exploiting the area around it.

\subsubsection{Dataset 2: Distributed Information}

Fig.~\ref{fig:results:matlab:b} shows a dataset created by randomly distributing rectangular blocks around the periphery of the map.
This dataset is representative of information being distributed around.
Fig.~\ref{fig:results:matlab:c} shows that the heuristic \AverageEntropy performs the best, while \algQvalAgg is able to match the heuristic. \RearSideVoxel saturates early on and performs worse. 
Fig.~\ref{fig:results:matlab:e} shows that \RearSideVoxel gets stuck exploiting an island of information. \AverageEntropy takes broader sweeps of the area thus gaining more information about the world. \algQvalAgg also shows a similar behavior of exploring the world map.

Thus we see that on changing the datasets the performance of the heuristics reverse while our data driven approach is able to adapt seamlessly.

\subsubsection{Other Datasets}

Fig.~\ref{fig:results:extra} shows results from other 2D datasets such as random disks and block worlds, where \algQvalAgg is able to outperform all heuristics. 

\subsection{3D Exploration}

We create a set of 3D exploration problems to test the algorithm on more realistic scenarios. The datasets comprises of 3D worlds created in Gazebo and simulated Kinect.
\subsubsection{Train on Synthetic, Test on Real}

To show the practical usage of our pipeline, we show a scenario where a policy is trained on synthetic data and tested on a real dataset. 

Fig.~\ref{fig:results:cpp:a} shows some sample worlds created in Gazebo to represent an office desk environment on which \algQvalAgg is trained. 
Fig.~\ref{fig:results:cpp:b} shows a dataset of an office desk collected by TUM Computer Vision Group \cite{sturm12iros}. The dataset is parsed to create a pair of pose and registered point cloud which can then be used to evaluate different algorithms.
Fig.~\ref{fig:results:cpp:c} shows that \algQvalAgg outperforms all heuristics. 
Fig.~\ref{fig:results:cpp:f} shows \algQvalAgg getting good coverage of the desk while the best heuristic \OcclusionAware misses out on the rear side of the desk.

\subsubsection{Other Datasets}

Fig.~\ref{fig:results:extra} shows more datasets where training and testing is done on synthetic worlds. 
\section{Conclusion}
\label{sec:conc}
We presented a novel data-driven imitation learning framework to solve budgeted information gathering problems. Our approach, \algQvalAgg, trains a policy to imitate a clairvoyant oracle that has full information about the world and can compute non-myopic plans to maximize information. The effectiveness of \algQvalAgg can be attributed to two main reasons: Firstly, as the distribution of worlds varies, the clairvoyant oracle is able to adapt and consequently \algQvalAgg adapts as well. Secondly, as the oracle computes non-myopic solutions, imitating it allows \algQvalAgg to also learn non-myopic behaviors.  





\section{Acknowledgement}
The authors thank Sankalp Arora for insightful discussions and open source code for exploration in MATLAB.

\end{document}